\documentclass[sigconf]{acmart}
\usepackage{tikz}
\usepackage{amsfonts}
\usepackage{amsmath}
\usepackage{booktabs}
\usepackage{threeparttable}
\usepackage{multirow}
\usepackage{siunitx}
\usepackage{booktabs}
\usepackage{graphicx}
\usepackage{bm}

\AtBeginDocument{%
  \providecommand\BibTeX{{%
    \normalfont B\kern-0.5em{\scshape i\kern-0.25em b}\kern-0.8em\TeX}}}

\setcopyright{acmcopyright}

\copyrightyear{2023}
\acmYear{2023}
\setcopyright{acmlicensed}\acmConference[KDD '23]{Proceedings of the 29th ACM SIGKDD Conference on Knowledge Discovery and Data Mining}{August 6--10, 2023}{Long Beach, CA, USA}
\acmBooktitle{Proceedings of the 29th ACM SIGKDD Conference on Knowledge Discovery and Data Mining (KDD '23), August 6--10, 2023, Long Beach, CA, USA}
\acmPrice{15.00}
\acmDOI{10.1145/3580305.3599430}
\acmISBN{979-8-4007-0103-0/23/08}

\acmConference[KDD '23]{the 29th ACM SIGKDD Conference on Knowledge Discovery and Data Mining}{August 06--10,
  2023}{Long Beach, CA}
\begin{document}

\title{MetricPrompt: Prompting Model as a Relevance Metric for Few-shot Text Classification}

\author{Hongyuan Dong}
\affiliation{%
  \institution{
  Research Center for Social Computing and Information Retrieval \\ 
  Harbin Institute of Technology}
  \city{Harbin}
  \state{Heilongjiang}
  \country{China}
}
\email{hydong@ir.hit.edu.cn}

\author{Weinan Zhang}
\affiliation{%
  \institution{
  Research Center for Social Computing and Information Retrieval \\ 
  Harbin Institute of Technology}
  \city{Harbin}
  \state{Heilongjiang}
  \country{China}
}
\email{wnzhang@ir.hit.edu.cn}

\author{Wanxiang Che}
\authornote{Corresponding author.}
\affiliation{%
  \institution{
  Research Center for Social Computing and Information Retrieval \\ 
  Harbin Institute of Technology}
  \city{Harbin}
  \state{Heilongjiang}
  \country{China}
}
\email{car@ir.hit.edu.cn}

\renewcommand{\shortauthors}{Hongyuan Dong, Weinan Zhang, \& Wanxiang Che}

\begin{abstract}
Prompting methods have shown impressive performance in a variety of text mining tasks and applications, especially few-shot ones.
Despite the promising prospects, the performance of prompting model largely depends on the design of prompt template and verbalizer.
In this work, we propose MetricPrompt, which eases verbalizer design difficulty by reformulating few-shot text classification task into text pair relevance estimation task.
MetricPrompt adopts prompting model as the relevance metric, further bridging the gap between Pre-trained Language Model's (PLM) pre-training objective and text classification task, making possible PLM's smooth adaption. 
Taking a training sample and a query one simultaneously, MetricPrompt captures cross-sample relevance information for accurate relevance estimation.
We conduct experiments on three widely used text classification datasets across four few-shot settings.
Results show that MetricPrompt outperforms manual verbalizer and other automatic verbalizer design methods across all few-shot settings, achieving new state-of-the-art (SOTA) performance. 
\end{abstract}






\maketitle

\section{Introduction}

Since unstructured text data takes up over 80\% information in our society, text mining is believed to have significant commercial value~\cite{korde2012text}.
Text classification is regarded as a fundamental and essential task in text mining, and the related techniques are used in various kinds of text mining applications, such as information retrieval~\cite{hoogeveen2018web, dwivedi2016automatic}, sentiment analysis~\cite{liu2012survey, pang2008opinion}, recommendation system~\cite{aggarwal2016content}, knowledge management~\cite{sumathy2013text}, document summarization~\cite{cao2017improving}, etc.
Recently proposed pre-trained language models (PLMs) achieve satisfactory text classification performance under data-rich setting~\cite{radford2018improving, radford2019language, devlin2019bert, raffel2019exploring, brown2020language}, but these models' Few-Shot Learning (FSL) ability still lags far behind human intelligence~\cite{brown2020language}. 


\begin{figure}[t]
\centering
\begin{tikzpicture}
\draw (0,0 ) node[inner sep=0] {\includegraphics[width=\columnwidth, trim={0cm 2.5cm 12.7cm -1cm}, clip]{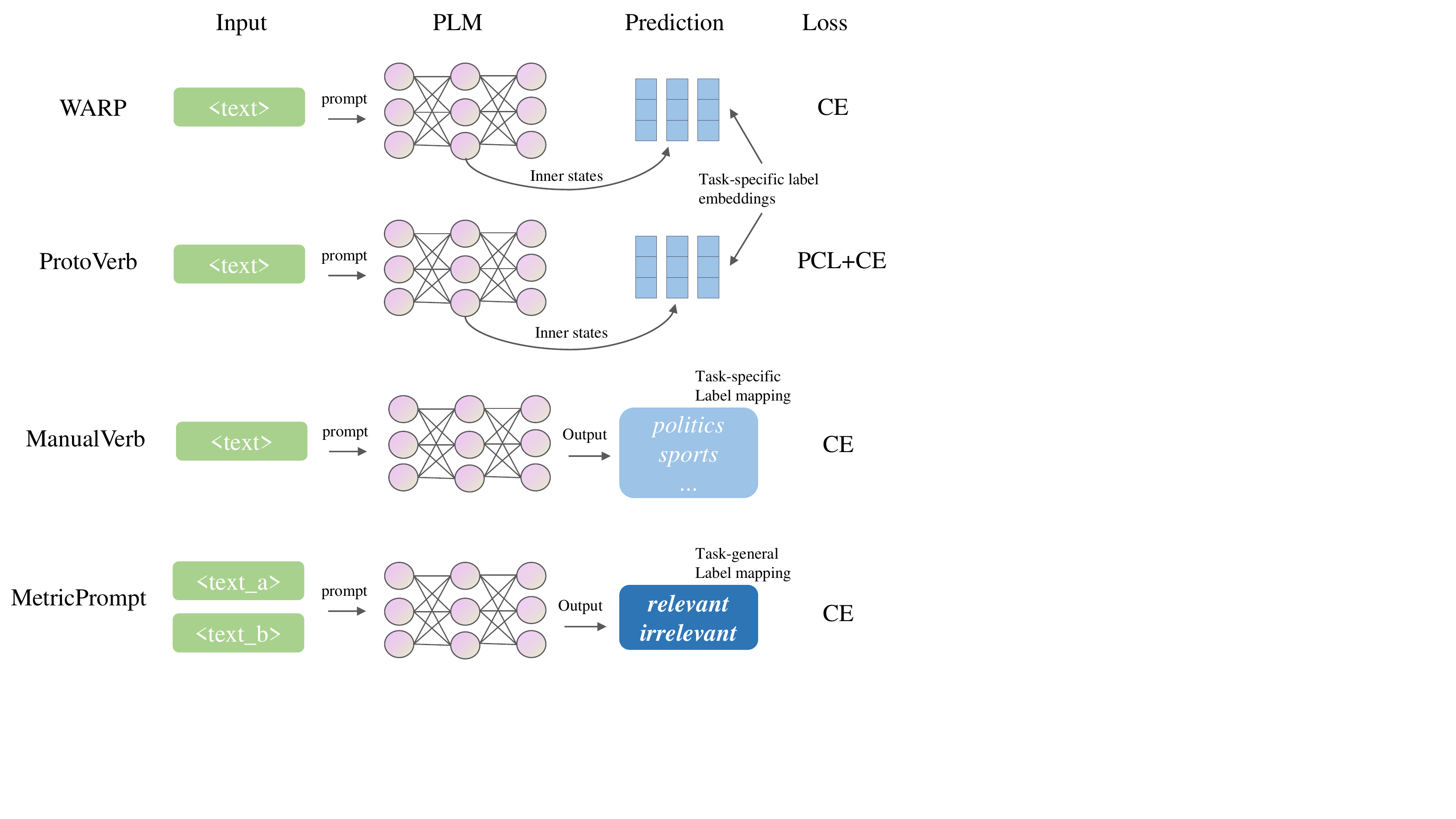}};
\end{tikzpicture}
\caption{
A comparison across verbalizer design methods. 
CE stands for Cross-Entropy loss and PCL is Prototypical Contrastive Learning loss~\cite{liprototypical}.
}\label{fig:intro}
\end{figure}


Prompting methods are proposed to better utilize PLM's general knowledge by aligning downstream tasks to its pre-training objective. 
Prompting method inserts sample text into prompt template to form prompted text. 
Prompting model takes as input the prompted text, and the result is obtained by projecting the model's output words to corresponding labels. 
Label mapping is conducted by \textit{verbalizer}, which serves as a critical part of the prompting model and determines its performance. 

\begin{figure*}[t]
\centering
\begin{tikzpicture}
\draw (0,0 ) node[inner sep=0] {\includegraphics[width=2\columnwidth, trim={0cm 9.5cm 6.7cm 0cm}, clip]{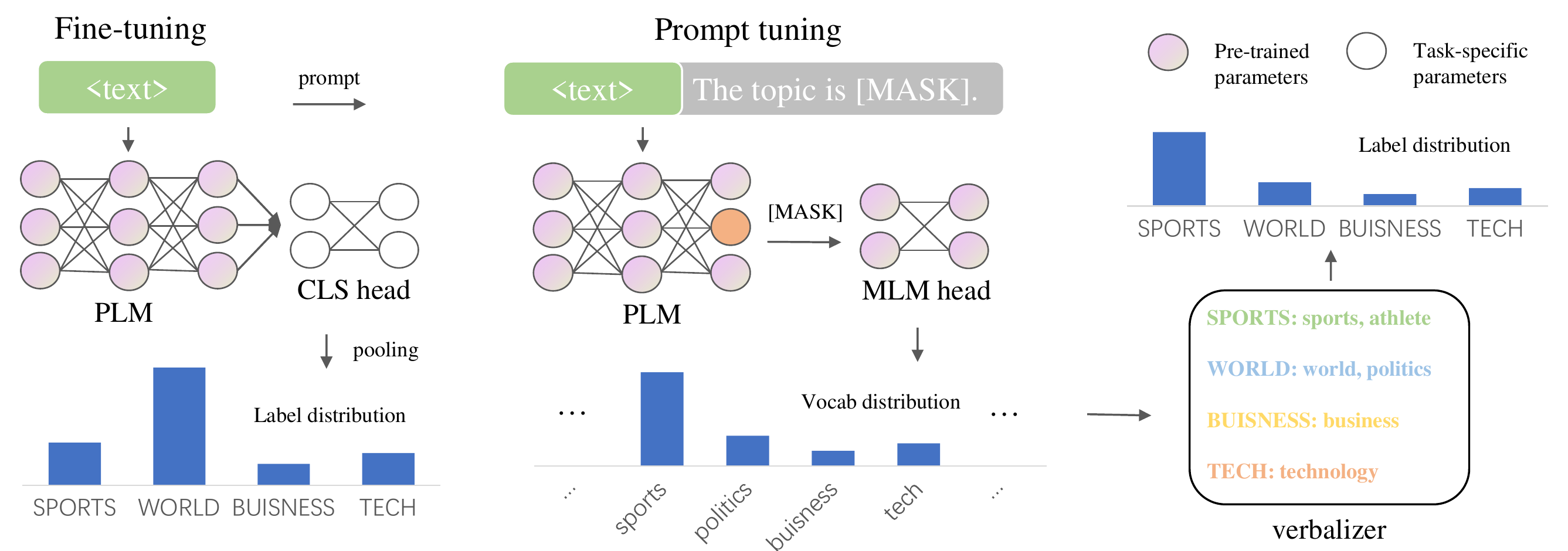}};
\end{tikzpicture}
\caption{
A comparison between vanilla fine-tuning and prompt tuning with MLM for text classification. 
}\label{fig:background}
\end{figure*}

Although achieving promising results in a wide variety of text mining tasks, prompting methods are very susceptible to sub-optimal verbalizer design~\cite{gao2020making}.
However, designing a proper verbalizer requires deep understanding of both downstream task and PLM's inner mechanism. 
In practice, this procedure can be extremely time and resource-consuming. 
To this end, automatic verbalizer design methods are proposed to ease the difficulty.
These algorithms can be classified into discrete verbalizer design and soft verbalizer design~\cite{liu2021pre} methods. 
Discrete verbalizer design methods, such as AVS~\cite{schick2021eacl}, LM-BFF~\cite{gao2020making} and AutoPrompt~\cite{shin2020autoprompt}, search each label's corresponding answer words in PLM's vocabulary to build the verbalizer. 
Soft verbalizer design methods like WARP~\cite{hambardzumyan2021warp} and ProtoVerb~\cite{cui2022prototypical} search for proper verbalizer parameters in an infinite continuous space and thus achieve better performance.

Despite the promising prospects of soft verbalizer, current methods' performance is still far from satisfactory. 
The main reason is that these methods' optimization and inference formulations are distinct from PLM's pre-training objective. 
As illustrated in Fig~\ref{fig:intro}, WARP~\cite{hambardzumyan2021warp} and ProtoVerb~\cite{cui2022prototypical} introduce task-specific label embeddings and use PLM's inner representation to make predictions. 
Although adopting prompting method, both methods force PLM to adapt to a distinct task formulation from its pre-training objective which only operates on output word probabilities.
What's worse, these label embeddings have to be trained from scratch in downstream tasks, leading to severe over-fitting problem.
As a result, PLM cannot adapt to downstream tasks smoothly. 


To tackle the above issues, we propose MetricPrompt, which frees human labor from task-specific verbalizer design by reformulating few-shot text classification task into text pair relevance estimation task. 
We re-organize training data into text pairs, and adopt prompting model to learn the relevance metric. 
The learned metric is then used to evaluate each query sample's relevance with training samples, and the classification result is obtained by pooling the estimated relevance scores.
As shown in Fig~\ref{fig:intro}, an explicit task-specific verbalizer is no longer required in our method.
Following PLM's pre-training objective, MetricPrompt only operates with PLM's output word probabilities, and thus enabling smooth adaption to downstream tasks. 
To produce accurate relevance estimations, MetricPrompt takes text pairs as input and aid estimation accuracy with cross-sample relevance information. 
Experiments on three widely used few-shot text classification datasets across four few-shot settings indicate that MetricPrompt achieves the highest few-shot text classification accuracy over previous SOTA verbalizer design baselines.

We summarize the contribution of this paper as below:

\textbf{(1)} We propose a novel prompting method MetricPrompt, which eases task-specific verbalizer design difficulty by reformulating few-shot classification task into relevance estimation problem and learning the relevance metric with prompting model.

\textbf{(2)} We conduct experiments on three widely used few-shot text classification datasets with four few-shot settings, and results show that MetricPrompt outperforms all automatic verbalizer baselines and even manual verbalizer which requires heavy human labor in task-specific verbalizer design.

\textbf{(3)} We provide analysis to demonstrate the extensibility and robustness of MetricPrompt, and explain its performance variance when equipped with different pooling methods.


All code and data will be publicly available at \href{https://github.com/Dousia/MetricPrompt}{https://github.com/\\Dousia/MetricPrompt}.

\begin{figure*}[t]
\centering
\begin{tikzpicture}
\draw (0,0 ) node[inner sep=0] {\includegraphics[width=2\columnwidth, trim={0cm 11cm 2cm 0cm}, clip]{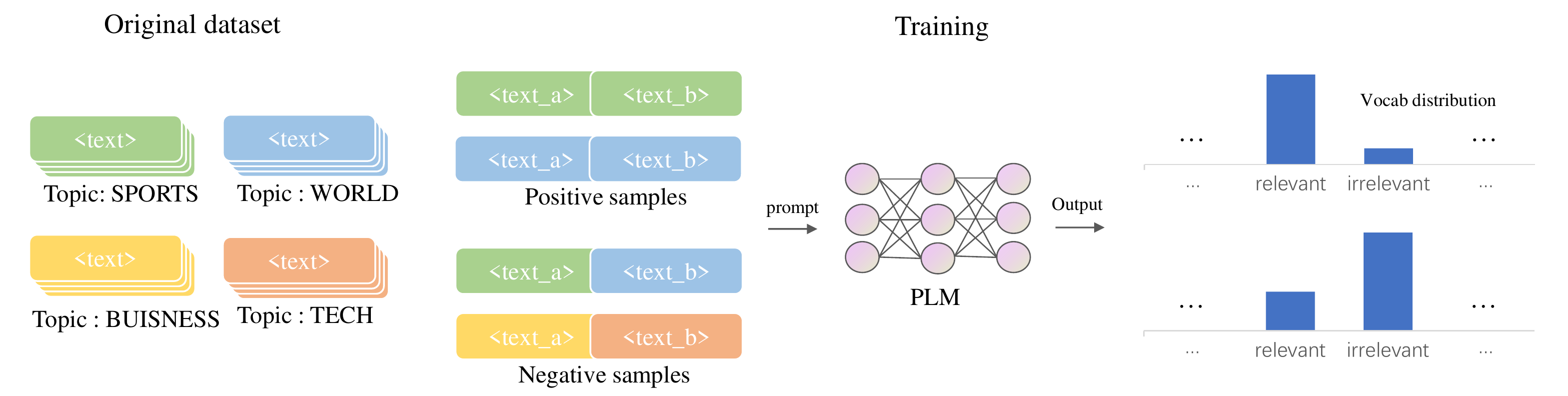}};
\end{tikzpicture}
\caption{
A demonstration of Metricprompt's data construction and training procedure. 
Original data is paired to form positive and negative samples, and the prompting model is optimized to map these samples to corresponding words. 
}\label{fig:method_1}
\end{figure*}

\section{Preliminaries and Related Work}

\subsection{Prompting methods}
Traditionally, PLM is adapted to downstream tasks via vanilla fine-tuning. 
As shown in Fig~\ref{fig:background}, fine-tuning pools PLM's last layer hidden states to form a sentence representation, and makes predictions with a task-specific classification head. 
Fine-tuning is effective given sufficient training data, but its performance degrades significantly under few-shot scenario. 
The newly initialized task-specific head is prone to over-fitting problem, and the gap between downstream task formulation and the pre-training objective hinders PLM's smooth adaption. 

Prompting methods are proposed to reformulate downstream tasks to enable PLM's smooth adaption to new tasks. 
A prompting model's pipeline is illustrated in Fig~\ref{fig:background}.
Denoting $p(\cdot)$ as the prompting function which fills the input text $\mathbf{x}$ into a prompt template, prompting model takes the prompted text and produces output word probability distribution over the vocabulary:
\begin{equation}
\label{eq: background_1}
f_{vocab}(p(\mathbf{x}); \theta) = P(p(\mathbf{x}); \theta), 
\end{equation}
where $P(\cdot)$ is a PLM parameterized by $\theta$. 
The output word probability is then mapped to final predicted probability distribution over classes with a verbalizer $v(\cdot)$:
\begin{equation}
\label{eq: background_2}
f_{cls}(p(\mathbf{x}); \theta) = v(f_{vocab}(p(\mathbf{x}); \theta)).  
\end{equation}

First proposed by~\citeauthor{radford2019language}, prompting methods have been used in a variety of text mining tasks, such as text classification~\cite{puri2019zero, schick2021eacl, PET, P-tuning}, text generation~\cite{schick2020few, li2021prefix, chen2021evaluating}, named entity recognition~\cite{cui2021template, hou2022inverse}, knowledge probing~\cite{petroni2019language, talmor2020olmpics}, etc. 
Prompting model alleviates the over-fitting problem under few-shot scenario significantly~\cite{brown2020language, PET, tam2021improving}. 
However, prompting model's performance relies largely on the selection of prompt template and verbalizer~\cite{liu2021pre}. 
Although a number of works focus on the design of prompt~\cite{gao2020making, shin2020autoprompt, li2021prefix, P-tuning}, less are proposed to ease verbalizer design difficulty. 
Verbalizer maps prompting model's output words to classification results directly, and therefore influences prompting model's performance significantly~\cite{gao2020making}.
In this work, we seek to free human labor from verbalizer design with less performance loss.

\subsection{Verbalizer Design Methods}

Current verbalizer design methods can be classified into manual verbalizer design, discrete verbalizer design and soft verbalizer design. 
A carefully hand-crafted verbalizer can achieve highly competitive performance in various text mining tasks~\cite{PET, schick2021eacl}. 
However, designing such a well-performing verbalizer relies heavily on human's accurate understanding of the target task and requires heavy trial-and-error work. 
Discrete verbalizer design methods frees human labor from tedious and time-consuming verbalizer design process. 
LM-BFF~\cite{gao2020making} generates proper answer words with large PLMs such as T5~\cite{raffel2019exploring}.
AutoPrompt~\cite{shin2020autoprompt} and AVS~\cite{schick2021eacl} initialize a verbalizer and optimize it to meet a predefined criteria iteratively. 
PETAL~\cite{schick2020coling} searches for answer words that maximize the likelihood of the training data.
These methods reduce human labor required by verbalizer design significantly, but their performance lags far behind carefully hand-crafted ones~\cite{cui2022prototypical}. 
To achieve better performance, soft verbalizer design methods render answer words from a fixed vocabulary as differentiable label embeddings represented in an infinite continuous space. 
Expanding the possibilities of verbalizer design space, these methods achieve better results than discrete verbalizer design algorithms~\cite{hambardzumyan2021warp, zhang2021differentiable, cui2022prototypical}. 

In this work, we ease task-specific verbalizer design difficulty by reformulating text classification task into text pair relevance estimation task.
In this way, no additional task-specific parameter is introduced, and the over-fitting problem is alleviated. 
Adopting prompting model as the relevance metric, PLM can adapt to new tasks more smoothly.
A recent work also views text classification task as natural language inference problem~\cite{plaza2022natural}, but it focuses on zero-shot scenario and hand-craft each label's description.
In comparison, our method tackle few-shot text classification tasks instead of zero-shot ones, and does not require human labor in task-specific verbalizer design.

\section{Methods}

In this section, we introduce the data construction, optimization and inference procedure of MetricPrompt in detail.

\subsection{Data construction}
Given a few-shot text classification dataset $\mathcal{D}$, we denote training data with $\mathcal{D}_{t}$ and query samples with $\mathcal{D}_{q}$. 
A sample is formulated as $d=(\mathbf{x}_d, \mathbf{y}_d)$, where $\mathbf{x}_d$ stands for sample text and $\mathbf{y}_d$ represents its label. 
Since MetricPrompt takes as input a pair of sample text, we construct training data as follows:
\begin{equation}
\label{eq: data_construction_train}
\mathcal{D}_{t}^{M} = \bigcup_{(d_i, d_j) \in \mathcal{D}_{t} \times \mathcal{D}_{t}} \{ (p(\mathbf{x}_{d_i}, \mathbf{x}_{d_j}), \mathbf{y}_{ij}) \},
\end{equation}
where $p(\cdot, \cdot)$ is MetricPromt's prompting function. It takes two pieces of sample text and produces a filled prompt template with the given text. 
We use ``$\mathbf{x}_{d_i}$ \texttt{[SEP]} A news of \texttt{[MASK]} topic: $\mathbf{x}_{d_j}$" as the prompt template. 
$\mathbf{y}_{ij}=\mathbb{I}(\mathbf{y}_{d_i}=\mathbf{y}_{d_j})$ indicates whether the pair of text are of the same class.

Similarly, we build query data for MetricPrompt as follows:
\begin{equation}
\label{eq: data_construction_query}
\mathcal{D}_{q}^{M} = \bigcup_{(d_i, d_j) \in \mathcal{D}_{q} \times \mathcal{D}_{t}} \{ (p(\mathbf{x}_{d_i}, \mathbf{x}_{d_j}), \mathbf{y}_{ij}) \}.
\end{equation}

\begin{figure*}[t]
\centering
\begin{tikzpicture}
\draw (0,0 ) node[inner sep=0] {\includegraphics[width=2\columnwidth, trim={0cm 1cm 2cm 9cm}, clip]{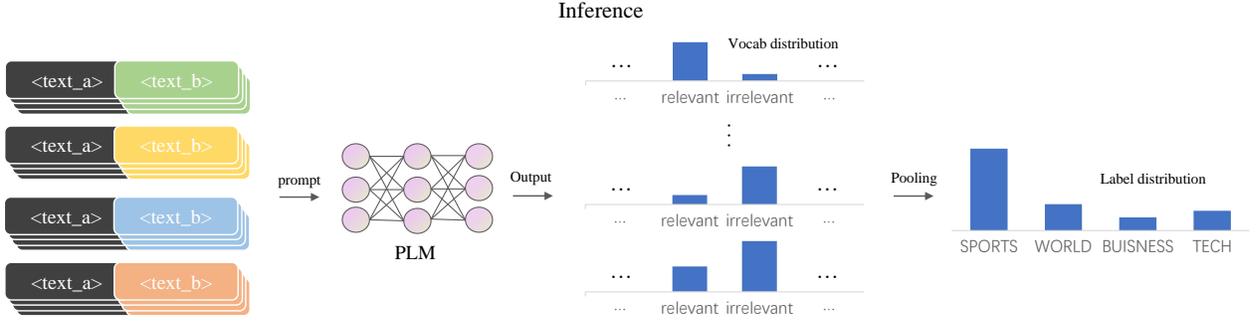}};
\end{tikzpicture}
\caption{
The inference procedure of MetricPrompt. 
A query sample is paired with each training sample, and the relevance scores are pooled to produce final classification probabilities.
}\label{fig:method_2}
\end{figure*}

The overall data construction process is illustrated in Fig~\ref{fig:method_1} and Fig~\ref{fig:method_2}.
We reorganize single samples of a given few-shot text classification task into paired samples, and the original multi-class classification problem is rendered as a relevance estimation task. 
No label's semantic representation is involved in this data construction procedure, and therefore MetricPrompt requires no human labor in task-specific verbalizer design.

\subsection{Optimization}
MetricPrompt differs from previous soft verbalizer design methods for we do not introduce task specific label embeddings, but instead reformulate few-shot text classification task as a text pair relevance estimation task to let loose the need of task-specific verbalizer.
By reformulating the task, MetricPrompt's optimization coincides with PLM's pre-training objectives.

Let $P(\cdot;\theta)$ be an MLM model parameterized by $\theta$ and $f_{vocab}(\cdot;\theta)$ be its output word probability over the vocabulary at \texttt{[MASK]} position. 
We define the optimization objective of MetricPrompt as follows:
\begin{equation}
\label{eq: optimization}
\begin{split}
\hat{\theta} 
&=\mathop{\arg\min}\limits_{\theta} \sum_{d^M \in \mathcal{D}_{t}^{M}} \mathcal{L}(f_{vocab}(\mathbf{x}_{d^M};\theta), \mathbf{y}_{d^M}) \\
&=\mathop{\arg\min}\limits_{\theta} \sum_{d^M \in \mathcal{D}_{t}^{M}} \mathcal{L}_{CE}(v(f_{vocab}(\mathbf{x}_{d^M};\theta)), \phi(\mathbf{y}_{d^M})), \\
&=\mathop{\arg\min}\limits_{\theta} \sum_{d^M \in \mathcal{D}_{t}^{M}} \mathcal{L}_{CE}(f_{cls}(\mathbf{x}_{d^M};\theta)), \phi(\mathbf{y}_{d^M})), 
\end{split}
\end{equation}
where $\phi(\cdot)$ stands for a probability distribution over label categories. 
The corresponding position of the input sample's label is set as 1 while others are set to be 0. 
$v(\cdot)$ represents a predefined task-general meta verbalizer, which projects output word probability $f_{vocab}(\cdot;\theta)$ to a binomial distribution $f_{cls}(\cdot;\theta)$. 
We task this meta verbalizer to aggregate logits at $\texttt{\{relevant}, \texttt{similar}, \texttt{consistent\}}$ to be the predicted logit of label 1, while logits at $\texttt{\{irrelevant}, \texttt{in-}$ $\texttt{consistent}, \texttt{different\}}$ are aggregated to be the logit of label 0.\footnote{Note that this meta verbalizer can be used in all few-shot text classification tasks, so MetricPrompt requires no extra human labor in task-specific verbalizer design.}
$\mathcal{L}$ is the loss function of MetricPrompt, which is defined as the cross-entropy loss between the probability distributions produced by the meta verbalizer $v(\cdot)$ and the ground truth distribution. 

MetricPrompt formulates few-shot text classification task's optimization objective to be a generalized MLM task, which is the minimization of cross-entropy loss between predicted word probability at \texttt{[MASK]} position and the ground truth one-hot distribution. The optimized prompting model can be used as a metric to estimate the relevance between two sample text. 


\subsection{Inference}
After optimization, the prompting model serves as a relevance metric during inference. 
As illustrated in Fig~\ref{fig:method_2}, we take an original query sample $d_q$ colored in black and pair it with all training samples colored differently to form inference samples.
Given an original training sample $d_i$, MetricPrompt computes its relevance score $s_{d_i}$ with $d_q$ as follows:
\begin{equation}
\label{eq: inference_1}
s_{d_i} = \Delta(f_{cls}(p(\mathbf{x}_{d_q}, \mathbf{x}_{d_i}); \hat{\theta})), 
\end{equation}
where $\Delta(\cdot)$ computes the difference between the binomial distribution's probability at 1 and 0. 
We denote $l$ as a label of the few-shot text classification task. 
Let $\mathcal{D}_{l}=\{d_i | d_i \in \mathcal{D}_{t}, \mathbf{y}_{d_i}=l_i\}$, MetricPrompt calculates $l$'s classification score $s_{l}$ by pooling its corresponding samples' relevance with $d_q$:
\begin{equation}
\label{eq: inference_2}
s_{l} 
= \sum_{d_i \in \mathcal{D}_{l}} s_{d_i} / |\mathcal{D}_{l}|. 
\end{equation}
Finally, MetricPrompt selects the label $\hat{l}$ with the highest the relevance score as the classification result:
\begin{equation}
\label{eq: inference_3}
\hat{l} 
= \mathop{\arg\max}\limits_{l} s_l. 
\end{equation}

We present MetricPrompt with sum pooling in Equation~\ref{eq: inference_2}.
This pooling function can also be replaced by max pooling and K-Nearest-Neighborhood (KNN)~\cite{cover1967nearest} pooling. 
Max pooling classifies query sample $d_q$ into its most relevant training sample's class. 
MetricPrompt works with max pooling by replacing Equation~\ref{eq: inference_2} with:
\begin{equation}
\label{eq: inference_4}
s_{l} 
= \max_{d_i \in \mathcal{D}_{l}} s_{d_i}. 
\end{equation}
For KNN pooling, we denote $\mathcal{D}_{topk}$ as $k$ training samples most relevant to $d_q$ in $\mathcal{D}_{t}$ and reformulate Equation~\ref{eq: inference_2} as:
\begin{equation}
\label{eq: inference_5}
s_{l} 
= |\{d_i | d_i \in \mathcal{D}_{topk}, \mathbf{y}_{d_i}=l_i\}|. 
\end{equation}
We set $k$ as half the size of training set. 
When several labels appear in $\mathcal{D}_{topk}$ for the same number of times, we select the sample obtaining the highest relevance score from these labels' corresponding samples, and classify $d_q$ to its class.

MetricPrompt renders prompting model as a relevance metric, and classify query samples according to its relevance with training samples of the few-shot task. 
Prompting model takes two pieces of sample text at once. 
Thus, MetricPrompt is able to use cross-sample information to estimate relevance and make predictions accurately.

\subsection{More Efficient Inference}
The cost of the inference procedure is relatively high because of the pairing strategy. 
To further improve the efficiency of MetricPrompt, we propose to use pivot samples to reduce the time complexity of the inference stage of MetricPrompt.

We propose to use the optimized prompting model to calculate the representativeness of each training sample. 
For a training sample $d_i$ labeled with $l$, we use $r_{d_i}$ to denote its representativeness, which is calculated as follows:
\begin{equation}
    r_{d_i} 
    = \frac{\sum_{d_j \in \mathcal{D}_l}{s_{d_j}}}{|\{d_j|d_j \in \mathcal{D}_l\}|} - \frac{\sum_{d_k \in \mathcal{D}_t-\mathcal{D}_l}{s_{d_k}}}{|\{d_k|d_k \in \mathcal{D}_t-\mathcal{D}_l\}|}. 
\end{equation}
$s_{d_j}$ represents the relevance score between samples $d_j$ and $d_i$. 
Based on this representativeness metric, we select the top $p$ samples with the highest representativeness scores from the training samples corresponding to each label.
These samples are marked as pivot samples.
During inference, we only pair each test sample with each label's pivot samples and compute their relevance scores to make classification prediction.

Pivot samples reduce the time complexity of MetricPrompt's inference process significantly.
Assume a few-shot text classification task with $n$ labels and $k$ samples per label.
Without introducing pivot samples, each test sample needs to be paired with $n*k$ training samples and compute relevance scores, resulting into a time complexity of $O(n*k)$. 
In contrast, prompting methods with human-designed or automatically designed verbalizer calculate the dot product similarity between the feature representations of test samples extracted by the pre-trained model and the feature representations of each label.
Since the total number of labels is $n$, the time complexity of these methods is only $O(n)$. 
By introducing pivot samples to improve the inference process, MetricPrompt only needs to estimate the relevance between each test sample and the pivot samples of each label, reducing the time complexity to $O(p*n)$. 
Because $p$ is a pre-defined constant, the time complexity of MetricPrompt's inference process accelerated with pivot samples is $O(n)$, which is consistent with other commonly used prompting methods. 
We set the number of pivot samples per label $p$ to 2 in the experiments.

\section{Experiments}
In this section, we first introduce the datasets used in our experiments. 
Then we describe our experiment settings and implementation details. 
Finally, we present experiment results and analyze MetricPrompt's performance.

\subsection{Datasets}
We conduct experiments with three widely used text classification tasks, which contain numerous classes and hence require extensive resources to build a proper manual verbalizer. 
We adopt AG’s News, Yahoo Answers topics~\cite{Zhang2015CharacterlevelCN} and DBPedia~\cite{lehmann2015dbpedia} as our text classification tasks.
The statistics of the datasets are given in Table ~\ref{tbl: datasets_statistics}.

\begin{table}[t]
\small
\begin{tabular}{c c c c}
\toprule
Dataset & \# Class & \# Test & Avg len
\\
\midrule
AG’s News & 4 & 7,600 & 52 \\
DBPedia & 14 & 70,000 & 68 \\
Yahoo & 10 & 60,000 & 130 \\
\bottomrule
\end{tabular}
\centering
\caption{
Statistics of the three text classification datasets used in our experiments. 
}
\label{tbl: datasets_statistics}
\vspace*{-3mm}
\end{table}

Since the average text length of the three datasets is short, we truncate all sample text to 120 tokens for better efficiency with little semantic meaning loss.

\subsection{Few-shot experiment settings}
We conduct experiments under 2, 4, 8 and 16-shot settings, where corresponding number of training samples are sampled from each dataset's training set randomly.
Models' performance are observed to fluctuate largely when the training set is sampled with different random seeds.  
Hence, we sample 10 training sets for each dataset and each few-shot setting to alleviate the influence of randomness in training set selection. 
All experiment results are reported as the average value of the model's performance on the 10 training sets. 

\begin{table}[t]
\small
\begin{tabular}{c c c c c}
\toprule
Dataset & 2-shot & 4-shot & 8-shot & 16-shot
\\
\midrule
AG’s News & 120 & 60 & 30 & 15 \\
DBPedia & 32 & 16 & 8 & 4 \\
Yahoo & 36 & 18 & 9 & 5 \\
\bottomrule
\end{tabular}
\centering
\caption{
The number of training epochs for given datasets and few-shot settings. 
}
\label{tbl: epochs}
\vspace*{-3mm}
\end{table}

\subsection{Implementation Details}
We implement MetricPrompt with PyTorch~\cite{paszke2019pytorch} and Huggingface~\cite{wolf2020transformers} framework, and baselines are implemented with OpenPrompt toolkit~\cite{ding2021openprompt}.
We introduce Kernl to accelerate the inference procedure.\footnote{\href{https://github.com/ELS-RD/kernl}{https://github.com/ELS-RD/kernl}}

We adopt BERT-base-uncased~\cite{devlin2019bert} as the backbone model for both MetricPrompt and all baseline models for fair comparison. 
Model parameters are optimized with AdamW optimizer~\cite{AdamW}, and the learning rate is set as 1e-5. 
We set total training steps proportionally to the size of training set, and the number of training epochs is adjusted accordingly. 
The size of training set varies across datasets and shot numbers, and the specific number of training epochs is given in Table~\ref{tbl: epochs}.

\begin{table*}[t]
\small
\begin{tabular}{c c c c}
\toprule
Method & Dataset & Prompt template & Task-specific verbalizer
\\
\midrule
\multirow{5}{*}{\textsc{ManualVerb}} & AG's News & A \texttt{[MASK]} news: \texttt{<text>} & sports, politics, business, technology \\
 & \multirow{2}{*}{DBPedia} & \multirow{2}{*}{\texttt{<text>} In this sentence, the topic is \texttt{[MASK]}} & company, school, artist, athlete, politics, transportation, \\
 & & & building, river, village, animal, plant, album, film, book \\
 & \multirow{2}{*}{Yahoo} & \multirow{2}{*}{A \texttt{[MASK]} question: \texttt{<text>}} & society, science, health, education, computers,  \\
 & & & sports, business, entertainment, relationships, politics \\
\midrule
\multirow{3}{*}{\textsc{AVS}} & AG's News & A \texttt{[MASK]} news: \texttt{<text>} & Automatically searched label words \\
 & DBPedia & \texttt{<text>} In this sentence, the topic is \texttt{[MASK]} & Automatically searched label words \\
 & Yahoo & A \texttt{[MASK]} question: \texttt{<text>} & Automatically searched label words \\
 \midrule
\multirow{3}{*}{\textsc{SoftVerb}} & AG's News & \multirow{3}{*}{\texttt{<text>} In this sentence, the topic is \texttt{[MASK]}} & \multirow{3}{*}{Soft label embeddings} \\
 & DBPedia &  &  \\
 & Yahoo &  &  \\
\midrule
\multirow{3}{*}{\textsc{ProtoVerb}} & AG's News & A \texttt{[MASK]} news: \texttt{<text>} & Soft label embeddings \\
 & DBPedia & \texttt{<text>} In this sentence, the topic is \texttt{[MASK]} & Soft label embeddings \\
 & Yahoo & A \texttt{[MASK]} question: \texttt{<text>} & Soft label embeddings \\
\midrule
\multirow{3}{*}{\textsc{MetricPrompt}} & AG's News & \multirow{3}{*}{\texttt{<text\_a>} A news of \texttt{[MASK]} topic: \texttt{<text\_b>}} & \multirow{3}{*}{$-$} \\
 & DBPedia &  &  \\
 & Yahoo &  &  \\
\bottomrule
\end{tabular}
\centering
\caption{
Prompt templates and task-specific verbalizers used by MetricPrompt and other baselines.
``-” means no task-specific verbalizer is required.
}
\label{tbl: 4_4}
\vspace*{-3mm}
\end{table*}

\subsection{Baselines}
We select several representative verbalizer design methods for comparison.
Among the listed baselines, only manual verbalizer requires human effort in task-specific verbalizer design. 

\textbf{Manual Verbalizer} (ManualVerb) uses hand-crafted verbalizer to map PLM's output word to classification labels. 

\textbf{Automatic Verbalize Search} (AVS)~\cite{schick2021eacl} is a search-based verbalizer design method. It initializes the verbalizer with random words, and improves answer words iteratively. 

\textbf{Soft Verbalizer} (SoftVerb)~\cite{hambardzumyan2021warp} represents each label with a trainable embedding. 
In the original work WARP, the prompt template is also represented as trainable embeddings.
In this work, however, we follow \citeauthor{cui2022prototypical} to use manual template for fair comparison.

\textbf{Prototypical Verbalizer} (ProtoVerb)~\cite{cui2022prototypical} also represents classification labels as soft embeddings and samples as features encoded by PLM.  
ProtoVerb adopts prototypical contrastive learning loss~\cite{liprototypical} instead of vanilla cross-entropy loss to optimize model parameters. 

We hand-craft a task-general prompt template and verbalizer for MetricPrompt.  
For other baselines, we make minor modifications to the default template and verbalizer used in OpenPrompt to unify the input format. 
An overview of prompt template and verbalizer used for our experiments is given in Table~\ref{tbl: 4_4}.

\subsection{Main Results}
\label{sec: 4_5}

\begin{table*}[t]
\begin{tabular}{l cc cc cc cc}
\toprule
\multicolumn{1}{c}{Method} &
\multicolumn{2}{c}{AG's News} &
\multicolumn{2}{c}{DBPedia} &
\multicolumn{2}{c}{Yahoo} &
\multicolumn{2}{c}{Average}
\\
\cmidrule(lr{0.5em}){2-3}\cmidrule(lr{0.5em}){4-5}\cmidrule(lr{0.5em}){6-7}\cmidrule(lr{0.5em}){8-9} 
& 2-shot & 4-shot & 2-shot & 4-shot & 2-shot & 4-shot & 2-shot & 4-shot 
\\
\midrule
\textsc{ManualVerb}       & $\textit{45.87}$ & $\textit{76.22}$ & $\textit{69.81}$ &  $\textit{84.15}$ & $\textit{34.82}$ &  $\textit{55.56}$ & $\textit{50.17}$ &  $\textit{71.98}$\\
\textsc{AVS}~\citeyearpar{schick2021eacl}        & $44.93$ & $57.49$ & $32.22$ &  $53.55$ & $21.88$ &  $28.44$ & $33.01$ &  $46.49$\\
\textsc{SoftVerb}~\citeyearpar{hambardzumyan2021warp}        & $48.86$ & $61.15$ & $53.98$ & $76.19$ & $22.63$ &  $33.43$ & $41.82$ & $56.92$\\
\textsc{ProtoVerb}~\citeyearpar{cui2022prototypical}        & $58.38$ & $65.04$ & $60.89$ & $74.49$ & $28.80$ &  $43.01$ & $49.36$ & $60.85$\\
\midrule
\textsc{$\text{MetricPrompt}_{\text{knn}}$}     & $62.69$ & $73.17$ & $66.27$ & $86.06$ & $26.02$ &  $50.90$ & $51.66$ &  $70.04$\\
\textsc{$\text{MetricPrompt}_{\text{max}}$}     & $65.64$ & $76.12$ & $\bm{71.28}$ & $\bm{88.44}$ & $\bm{28.85}$ &  $52.99$ & $\bm{55.26}$ &  $72.52$\\
\textsc{$\text{MetricPrompt}_{\text{mean}}$}       & $\bm{65.77}$ & $\bm{76.33}$ & $71.20$ & $\bm{88.44}$ & $28.76$ &  $\bm{53.54}$ & $55.24$ &  $\bm{72.77}$ \\
\textsc{$\text{MetricPrompt}_{\text{pivot}}$}       & $65.76$ & $74.53$ & $71.20$ & $86.12$ & $28.76$ &  $51.32$ & $55.24$ &  $70.66$ \\
\bottomrule
\end{tabular}
\centering
\caption{
Experiment results in terms of accuracy under 2-shot and 4-shot settings. Italic score means human labor is involved in task-specific verbalizer design, and bold number indicates the best result among methods requiring no human labor. 
}
\label{tbl: main_24}
\vspace*{-3mm}
\end{table*}

\begin{table*}[t]
\begin{tabular}{l cc cc cc cc}
\toprule
\multicolumn{1}{c}{Method} &
\multicolumn{2}{c}{AG's News} &
\multicolumn{2}{c}{DBPedia} &
\multicolumn{2}{c}{Yahoo} &
\multicolumn{2}{c}{Average}
\\ 
\cmidrule(lr{0.5em}){2-3}\cmidrule(lr{0.5em}){4-5}\cmidrule(lr{0.5em}){6-7}\cmidrule(lr{0.5em}){8-9} 
& 8-shot & 16-shot & 8-shot & 16-shot & 8-shot & 16-shot & 8-shot & 16-shot 
\\
\midrule
\textsc{ManualVerb}       & $\textit{78.94}$ & $\textit{83.66}$ & $\textit{94.24}$ & $\textit{97.27}$ & $\textit{58.30}$ &  $\textit{62.42}$ & $\textit{77.16}$ & $\textit{81.12}$\\
\textsc{AVS}~\citeyearpar{schick2021eacl}      & $71.37$ & $77.81$ & $75.91$ & $85.36$ & $46.53$ & $57.68$ & $64.60$ &  $73.62$\\
\textsc{SoftVerb}~\citeyearpar{hambardzumyan2021warp}       & $73.28$ & $80.61$ & $90.34$ & $96.93$ & $45.01$ &  $59.09$ & $69.54$ &  $78.88$\\
\textsc{ProtoVerb}~\citeyearpar{cui2022prototypical}       & $75.57$ & $80.31$ & $87.45$ &  $\bm{97.16}$ & $52.87$ &  $61.57$ & $71.96$ &  $79.68$\\
\midrule
\textsc{$\text{MetricPrompt}_{\text{knn}}$}     & $80.64$ & $84.43$ & $94.25$ & $96.55$ & $58.09$ &  $62.05$ & $77.66$ & $81.01$\\
\textsc{$\text{MetricPrompt}_{\text{max}}$}     & $81.03$ & $84.27$ & $94.28$ & $96.55$ & $59.68$ &  $62.66$ & $78.33$ & $81.16$\\
\textsc{$\text{MetricPrompt}_{\text{mean}}$}       &  $\bm{82.04}$ & $\bm{84.69}$ & $\bm{94.57}$ & $96.59$ & $\bm{59.68}$ & $\bm{62.45}$ & $\bm{78.76}$ & $\bm{81.24}$\\
\textsc{$\text{MetricPrompt}_{\text{pivot}}$}       & $81.19$ & $84.15$ & $94.13$ & $96.22$ & $58.63$ &  $61.78$ & $77.98$ &  $80.72$ \\
\bottomrule
\end{tabular}
\centering
\caption{
Experiment results in terms of accuracy under 8-shot and 16-shot settings. Italic score means human labor is involved in task-specific verbalizer design, and bold number indicates the best result among methods requiring no human labor. 
}
\label{tbl: main_816}
\vspace*{-3mm}
\end{table*}


We conduct experiments on three text classification datasets with different text styles under four few-shot settings. 
Experiment results for 2 and 4-shot settings are listed in Table~\ref{tbl: main_24}, while 8 and 16-shot's experiment results are shown in Table~\ref{tbl: main_816}. 
MetricPrompt outperforms previous SOTA automatic verbalizer design method ProtoVerb by a large margin, improving 2-shot accuracy by $5.88$ ($11.91\%\uparrow$), 4-shot accuracy by $11.92$ ($19.59\%\uparrow$), 8-shot accuracy by $6.80$ ($9.45\%\uparrow$) and 16-shot accuracy by $1.56$ ($1.96\%\uparrow$). 
MetricPrompt's performance even surpasses that of ManualVerb under all few-shot settings without any human labor involved in task-specific verbalizer design. 

Meanwhile, we have following observations:

\textbf{(1)} MetricPrompt is the only prompting method outperforming ManualVerb without any human labor involved in task-specific verbalizer design. 
MetricPrompt benefits from paired input text, which enables the model to use cross-sample information to make up the lack of extra human knowledge and trial-and-error work.

\textbf{(2)} MetricPrompt achieves the highest score over automatic verbalizer design methods. 
Compared with AVS, MetricPrompt does not restrict each class's representation to several sub-optimal words from the vocabulary, but instead represent it with corresponding training samples, leading to more accurate semantic representation. 
For the comparison with SoftVerb and ProtoVerb, we attribute MetricPrompt's leading performance to the smaller gap between its task formulation and PLM's pre-training objective. 
Unlike SoftVerb and ProtoVerb, MetricPrompt does not operate on PLM's inner representations, but functions with only the output word probability distribution at \texttt{[MASK]} position, enabling PLM to adapt to few-shot text classification task more smoothly.
Moreover, MetricPrompt does not introduce task-specific parameters to be trained from scratch, avoiding over-fitting problem under few-shot scenarios. 

\textbf{(3)} MetricPrompt achieves comparable results with mean pooling and max pooling, but a performance drop is witnessed with KNN pooling.  
We ascribe the performance gap to the incompatibility between KNN pooling and the non-uniform distribution of MetricPrompt's predicted relevance scores.
We further discuss this phenomenon in Section~\ref{sec: 5_3}.

\section{Analysis}
In this section, we further evaluate MetricPrompt from different aspects to illustrate its effectiveness. 

\begin{table*}[t]
\begin{tabular}{l ccc l ccc l ccc}
\toprule
\multicolumn{1}{c}{Method} &
\multicolumn{3}{c}{AG's News} &
\multicolumn{1}{c}{Method} &
\multicolumn{3}{c}{DBPedia} &
\multicolumn{1}{c}{Method} &
\multicolumn{3}{c}{Yahoo}
\\
\cmidrule(lr{0.5em}){2-4}\cmidrule(lr{0.5em}){6-8}\cmidrule(lr{0.5em}){10-12}
& 1-shot & 2-shot & 4-shot & & 1-shot & 2-shot & 4-shot & & 1-shot & 2-shot & 4-shot
\\
\midrule
\textsc{ProtoVerb}     & $46.79$ & $58.38$ & $65.04$ & \textsc{ProtoVerb} & $45.86$ & $60.89$ &  $74.49$ & \textsc{ProtoVerb} & $21.60$ & $28.80$ & $43.01$ \\
\quad\textsc{+DBPedia}     & $57.43$ & $65.72$ & $71.27$ & \quad\textsc{+AG's News} & $49.00$ & $63.29$ &  $75.56$ & \quad\textsc{+AG's News} & $28.93$ & $39.15$ & $49.96$ \\
\quad\textsc{+Yahoo}       &  $63.63$ & $71.84$ & $75.34$ & \quad\textsc{+Yahoo} & $\bm{\underline{54.33}}$ & $66.78$ &  $77.56$ & \quad\textsc{+DBPedia} & $30.13$ & $39.57$ & $51.39$ \\
\midrule
\textsc{MetricPrompt}     & $39.16$ & $65.77$ & $76.33$ & \textsc{MetricPrompt} & $32.31$ & $71.20$ & $88.44$ & \textsc{MetricPrompt} & $18.80$ & $28.76$ & $53.54$ \\
\quad\textsc{+DBPedia}     & $\underline{66.95}$ & $\underline{71.40}$ & $\underline{77.34}$ & \quad\textsc{+AG's News} & $\underline{53.23}$ & $\underline{74.21}$ & $\underline{88.34}$ & \quad\textsc{+AG's News} & $\underline{32.10}$ & $\bm{\underline{44.27}}$ & $\underline{52.29}$ \\
\quad\textsc{+Yahoo}       &  $\bm{\underline{71.00}}$ & $\bm{\underline{73.99}}$ & $\bm{\underline{79.57}}$ & \quad\textsc{+Yahoo} & $53.03$ & $\bm{\underline{76.41}}$ &  $\bm{\underline{89.47}}$ & \quad\textsc{+DBPedia} & $\bm{\underline{32.77}}$ & $\underline{43.63}$ & $\bm{\underline{53.78}}$ \\
\bottomrule
\end{tabular}
\centering
\caption{
The performance of MetricPrompt and ProtoVerb with additional OOD training data. 
Underlined number indicates the best result under the same few-shot and OOD setting. 
Bold number represents the best result on the few-shot task. 
}
\label{tbl: extensibility}
\vspace*{-3mm}
\end{table*}

\subsection{Extensibility with Out-Of-Domain Data}
\label{sec: 5_1}
Since it is impractical to always obtain sufficient training data for downstream few-shot text classification tasks, we further evaluate MetricPrompt's extensibility with Out-Of-Domain (OOD) data.
We use 16-shot training sets of other datasets to aid each task's 1, 2 and 4-shot classification, where models' performance remains to be improved. 
We adopt mean pooling and simply mix up training samples of the original dataset and OOD ones to train the model. 
For ProtoVerb baseline, we first train the model on OOD training data, and then re-initialize prototype parameters for the training on the original dataset. 

As shown in Table~\ref{tbl: extensibility}, MetricPrompt achieves much higher accuracy when boosted by OOD training data. 
Compared with previous SOTA baseline ProtoVerb, MetricPrompt obtains better prediction accuracy in 17 out of 18 few-shot and OOD data settings (underlined numbers in the table), showing remarkable extensibility with OOD samples. 
Although OOD data improves the model's performance under most settings, AG's News training set fail to aid DBPedia 4-shot and Yahoo 4-shot performance.
We ascribe this failure to the abundance of training samples under these settings. 
DBPedia and Yahoo contains 4096 and 1600 training samples under 4-shot setting, which is sufficient for the model to adapt to the relevance estimation task. 
OOD data serves more as noises and therefore harms the model's performance. 

\begin{table*}[t]
\begin{tabular}{l cc cc cc cc}
\toprule
\multicolumn{1}{c}{Method} &
\multicolumn{2}{c}{1 wrong} &
\multicolumn{2}{c}{2 wrong} &
\multicolumn{2}{c}{4 wrong} &
\multicolumn{2}{c}{Average}
\\
\cmidrule(lr{0.5em}){2-3}\cmidrule(lr{0.5em}){4-5}\cmidrule(lr{0.5em}){6-7}\cmidrule(lr{0.5em}){8-9}
& 8-shot & 16-shot & 8-shot & 16-shot & 8-shot & 16-shot & 8-shot & 16-shot \\
\midrule
\textsc{ProtoVerb}     & $2.79$ & $0.83$ & $4.95$ & $1.85$ & $11.31$ &  $3.71$ & $6.35$ & $2.13$ \\
\textsc{$\text{MetricPrompt}_{\text{knn}}$}     & $5.74$ & $2.61$ & $5.66$ & $3.08$ & $12.38$ &  $3.38$ & $7.93$ & $3.02$ \\
\textsc{$\text{MetricPrompt}_{\text{max}}$}     & $\bm{1.59}$ & $0.59$ & $3.12$ & $\bm{0.89}$ & $\bm{7.01}$ &  $\bm{1.21}$ & $3.91$ & $\bm{0.90}$ \\
\textsc{$\text{MetricPrompt}_{\text{mean}}$}     & $1.81$ & $\bm{0.55}$ & $\bm{2.72}$ & $1.04$ & $7.06$ & $1.52$ & $\bm{3.86}$ & $1.04$ \\
\bottomrule
\end{tabular}
\centering
\caption{
Models' performance drop under AG's News 8 and 16-shot settings with 1, 2 and 4 noisy samples.
Bold number indicates the least drop among all methods.
}
\label{tbl: robustness}
\vspace*{-3mm}
\end{table*}

It is worth noticing that MetricPrompt's performance improvement under 1-shot setting is abnormally high. 
MetricPrompt underperforms ProtoVerb on the three datasets without OOD data, because MetricPrompt only takes two identical pieces of text as positive sample under 1-shot setting, leading to severe over-fitting problem. 
The model is optimized to only produce high relevance score when given two identical pieces of text. 
Diversified OOD data alleviate the over-fitting problem effectively, and therefore improve MetricPrompt's 1-shot performance by a large fraction. 

\subsection{Robustness against Noisy Samples}
Noisy samples harm few-shot text classification model's performance severely due to the lack of supervision signal. 
In this section, we evaluate MetricPrompt's robustness against noisy samples on AG's News dataset. 
Following~\citeauthor{cui2022prototypical}, we conduct experiments under 8 and 16-shot settings for training stability.
We replace 1, 2 and 4 training samples' labels randomly to introduce noises, and evaluate MetricPrompt's performance when equipped with mean, max and KNN pooling. 
We compare our method with previous SOTA baseline ProtoVerb, which shows the best robustness against noisy samples among automatic verbalizer design methods~\cite{cui2022prototypical}. 
The performance drop caused by noisy samples is displayed in Table~\ref{tbl: robustness}. 

Compared with ProtoVerb, MetricPrompt suffers from less performance drop and achieves higher classification accuracy with mean and max pooling, while KNN pooling leads to worse performance. 
We attribute the large performance drop of KNN pooling to its susceptibility to the variance of each class's training sample number, which is introduced by noisy samples.
We provide a detailed analysis in Section~\ref{sec: 5_3}.

\subsection{Comparison across Pooling Methods}
\label{sec: 5_3}
In this part, we analyze different pooling methods with clean and noised training data to explain their performance gap. 

\begin{figure}[t]
\centering
\begin{tikzpicture}
\draw (0,0 ) node[inner sep=0] {\includegraphics[width=\columnwidth, trim={0cm 0cm 0cm 1cm}, clip]{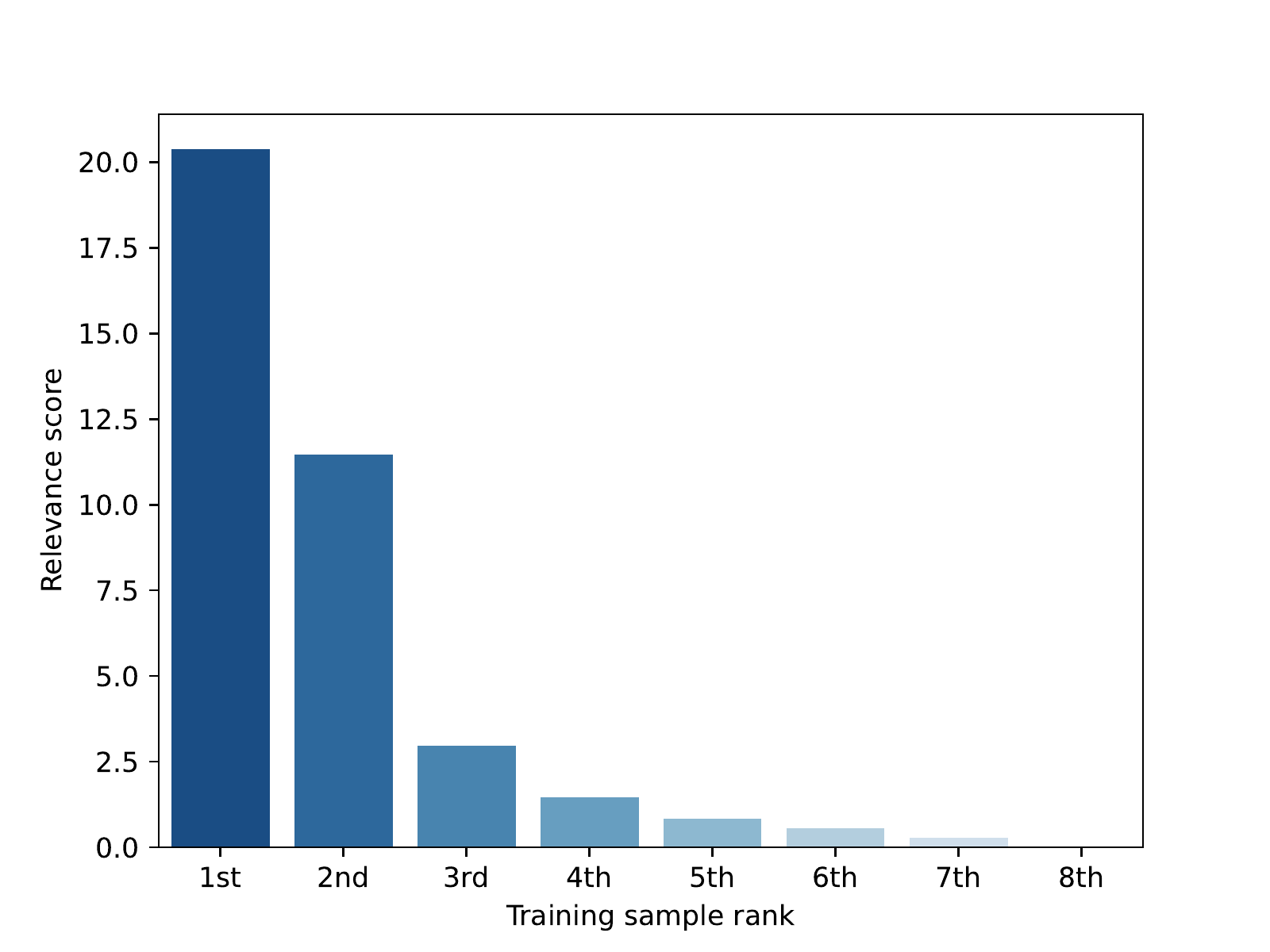}};
\end{tikzpicture}
\caption{
Average relevance scores between each query sample and all training samples under AG's News 2-shot setting. 
The scores are sorted and shifted to non-negative region.
}\label{fig: 5_3_1}
\end{figure}

Firstly, we focus on clean data scenario without noisy samples.
We choose AG's News 2-shot setting and compute the average relevance score of each query sample's most relevant training sample, second most relevant training sample, etc. 
As shown in Fig~\ref{fig: 5_3_1}, the distribution of relevance scores is highly non-uniform. 
The highest relevance score is much larger than others, so the max relevance score serves as a decisive factor for MetricPrompt with mean pooling.
Therefore, mean pooling shows similar behavior with max pooling. 
KNN pooling, however, adopts voting strategy which ignores score value information, leading to deteriorated performance.

\begin{figure*}[t]
\centering
\begin{tikzpicture}
\draw (0,0 ) node[inner sep=0] {\includegraphics[width=2\columnwidth, trim={0cm 0cm 0cm 0cm}, clip]{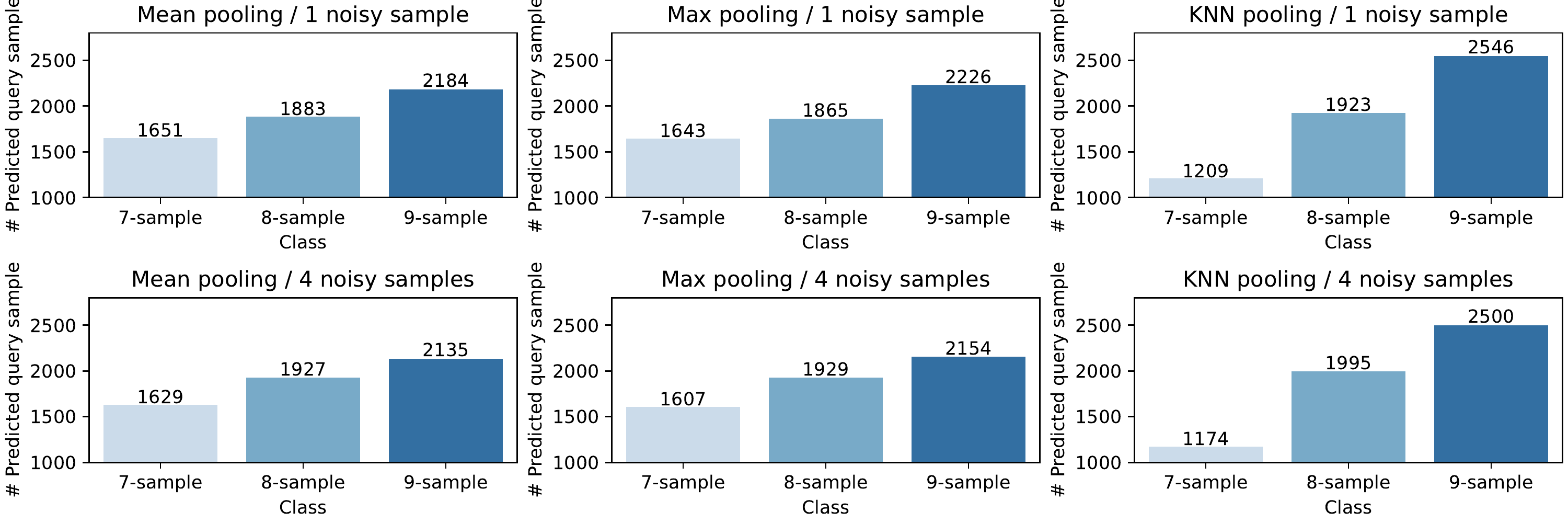}};
\end{tikzpicture}
\caption{
Average query sample number classified to classes with 7, 8 and 9 training samples under AG's News 8-shot setting. 
``\# Predicted query sample” indicates the average number of query samples predicted to the class. 
KNN pooling shows stronger preference to 9-sample classes than classes with fewer training samples. 
}\label{fig: 5_3_2}
\end{figure*}


We then analyze MetricPrompt's performance when noisy samples are introduced.
We first categorize classes of AG's News dataset 8-shot training sets according to the number of their corresponding training samples.
Then we collect the statistics of the average predicted query sample number for each type of class and show them in Figure~\ref{fig: 5_3_2}.
KNN pooling shows significantly stronger preference to classes with more training samples than mean pooling and max pooling do. 
Since the distribution of MetricPrompt's relevance score except the top ones is relatively even, samples from classes with large training set are more likely to become the majority of KNN pooling's top $k$ relevant samples.
Without considering relevance score value, a large drop in KNN pooling's performance is witnessed.
On the contrary, mean pooling and max pooling take each training sample's relevance score value into consideration, so the influence of training sample number is mitigated. 
As a result, they suffer from smaller performance drops than KNN pooling do. 


\subsection{Influence of Pivot Sample Number}

\begin{table*}[t]
    \begin{tabular}{l cc cc cc cc}
    \toprule
    \multicolumn{1}{c}{Method} &
    \multicolumn{2}{c}{AG's News} &
    \multicolumn{2}{c}{DBPedia} &
    \multicolumn{2}{c}{Yahoo} &
    \multicolumn{2}{c}{Average}
    \\
    \cmidrule(lr{0.5em}){2-3}\cmidrule(lr{0.5em}){4-5}\cmidrule(lr{0.5em}){6-7}\cmidrule(lr{0.5em}){8-9} 
    & 2-shot & 4-shot & 2-shot & 4-shot & 2-shot & 4-shot & 2-shot & 4-shot 
    \\
    \midrule
    \textsc{ProtoVerb}        & $58.38$ & $65.04$ & $60.89$ & $74.49$ & $28.80$ &  $43.01$ & $49.36$ & $60.85$\\
    \textsc{$\text{MetricPrompt}_{\text{1pivot}}$}       & $61.77$ & $71.41$ & $65.01$ & $84.07$ & $25.92$ &  $48.42$ & $50.90$ &  $67.97$ \\
    \textsc{$\text{MetricPrompt}_{\text{2pivot}}$}       & $\bm{65.76}$ & $74.53$ & $\bm{71.20}$ & $86.12$ & $\bm{28.76}$ &  $51.32$ & $\bm{55.24}$ &  $70.66$ \\
    \textsc{$\text{MetricPrompt}_{\text{4pivot}}$}       & $\bm{65.76}$ & $\bm{76.33}$ & $\bm{71.20}$ & $\bm{88.44}$ & $\bm{28.76}$ &  $\bm{53.54}$ & $\bm{55.24}$ &  $\bm{72.77}$ \\
    \bottomrule
    \end{tabular}
    \centering
    \caption{
    Experiment results with pivot samples under 2-shot and 4-shot settings. 
    Bold number indicates the best result. 
    }
    \label{tbl: pivot_24}
    \vspace*{-3mm}
\end{table*}

\begin{table*}[t]
    \begin{tabular}{l cc cc cc cc}
    \toprule
    \multicolumn{1}{c}{Method} &
    \multicolumn{2}{c}{AG's News} &
    \multicolumn{2}{c}{DBPedia} &
    \multicolumn{2}{c}{Yahoo} &
    \multicolumn{2}{c}{Average}
    \\ 
    \cmidrule(lr{0.5em}){2-3}\cmidrule(lr{0.5em}){4-5}\cmidrule(lr{0.5em}){6-7}\cmidrule(lr{0.5em}){8-9} 
    & 8-shot & 16-shot & 8-shot & 16-shot & 8-shot & 16-shot & 8-shot & 16-shot 
    \\
    \midrule
    \textsc{ProtoVerb}       & $75.57$ & $80.31$ & $87.45$ &  $\bm{97.16}$ & $52.87$ &  $61.57$ & $71.96$ &  $79.68$\\
    \textsc{$\text{MetricPrompt}_{\text{1pivot}}$}       & $79.23$ & $84.17$ & $93.38$ & $96.04$ & $57.20$ &  $61.15$ & $76.60$ &  $80.45$ \\
    \textsc{$\text{MetricPrompt}_{\text{2pivot}}$}       & $\bm{81.19}$ & $84.15$ & $94.13$ & $96.22$ & $58.63$ &  $61.78$ & $77.98$ &  $80.72$ \\
    \textsc{$\text{MetricPrompt}_{\text{4pivot}}$}       & $81.13$ & $\bm{84.62}$ & $\bm{94.42}$ & $96.49$ & $\bm{59.21}$ &  $\bm{61.93}$ & $\bm{78.25}$ &  $\bm{81.01}$ \\
    \bottomrule
    \end{tabular}
    \centering
    \caption{
    Experiment results with pivot samples under 8-shot and 16-shot settings. 
    Bold number indicates the best result. 
    }
    \label{tbl: pivot_816}
    \vspace*{-3mm}
\end{table*}

In this part, we investigate the influence of pivot sample numbers to the performance of MetricPrompt. 
We conduct experiments on the three datasets across four few-shot settings with pivot sample number $p$ set as 1, 2 and 4, the performance of MetricPrompt under different few-shot settings are displayed Table~\ref{tbl: pivot_24} and Table~\ref{tbl: pivot_816}. 

As shown in the tables, the performance of MetricPrompt correlates with the number of pivot samples positively. 
As the number of pivot samples increase, MetricPrompt captures each label's semantic meaning more accurately and therefore achieves better performance. 
It is worth noticing that even if only one pivot sample is selected for each class, MetricPrompt still outperforms ProtoVerb under the four few-shot settings. 
The selection of pivot sample number $p$ serves as a trade-off between classification accuracy and efficiency.
In real world applications, MetricPrompt can adapt to varying scenarios with different requirements by adjusting the number of pivot samples. 


\section{Conclusions and Future Work}
In this work, we propose MetricPrompt, which frees human labor from task-specific verbalizer design by reformulating few-shot text classification task into a text pair relevance estimation problem. 
MetricPrompt prompts query-training text pair to fulfill relevance estimation, which coincides with PLM's pre-training objective and thus enables smooth adaption to downstream tasks.
Taking a pair of sample text simultaneously, MetricPrompt introduces cross-sample information for better accuracy.
Experiments on three widely used text classification datasets under four few-shot settings indicate MetricPrompt's leading performance over previous SOTA baselines. 
Our analysis further demonstrates MetircPrompt's promising extensibility and robustness, and explains its performance variance with different pooling methods and pivot sample numbers. 

Although MetricPrompt achieves satisfactory few-shot text classification performance in our experiments, its performance with large language models remains to be explored. 
Current large language models achieve impressive performance in few-shot text classification tasks with prompting methods, but they still suffer from prompting methods' susceptibility to the design of verbalizers. 
When given classes which are difficult to be described with several words, these models' performance deteriorates significantly. 
The proposed MetricPrompt is not bounded with specific backbone model, and can be easily generalized to large language models.
We look forward to using MetricPrompt to ease human effort from verbalizer design and further improve the few-shot text classification accuracy of large language models. 

\section*{Acknowledgements}
This work was supported by the National Key R\&D Program of China via grant 2020AAA0106501 and the National Natural Science Foundation of China (NSFC) via grant 62236004 and 61976072.

\bibliographystyle{ACM-Reference-Format}
\bibliography{sample-base}

\end{document}